\documentclass[11pt]{article}

\usepackage[final]{acl}

\usepackage{times}
\usepackage{latexsym}

\usepackage[T1]{fontenc}

\usepackage[utf8]{inputenc}
\usepackage{amssymb}

\usepackage{microtype}

\usepackage{inconsolata}

\usepackage{graphicx}
\usepackage{enumitem}
\usepackage{booktabs}  
\usepackage{multirow}
\usepackage{pifont}
\usepackage{xspace}
\usepackage{listings}
\usepackage{xcolor}
\usepackage{colortbl}
\usepackage{nicematrix}
\usepackage{afterpage}
\usepackage{textcomp}

\usepackage{algorithm}
\usepackage[noend]{algpseudocode}
\usepackage{amsmath}
\usepackage{bbm}
\usepackage{comment}

\definecolor{LightCyan}{rgb}{0.88,1,1}
\newcolumntype{a}{>{\columncolor{LightCyan}}c}
\definecolor{myred}{rgb}{0.6, 0, 0}
\newcommand\xx[1]{\textcolor{black}{{#1}}}

\hypersetup{breaklinks=true}

\title{Verbal-R3: Verbal Reranker as the\\Missing Bridge between Retrieval and Reasoning}

\author{
    Sangkwon Park\textsuperscript{\textnormal{1}}\thanks{~~Equal Contribution},
    Donghun Kang\textsuperscript{\textnormal{1}}\footnotemark[1],
    Jisoo Mok\textsuperscript{\textnormal{4}}\thanks{~~Corresponding Authors},
    Sungroh Yoon\textsuperscript{\textnormal{1,2,3}}\footnotemark[2] \\ \\
\textsuperscript{\textnormal{1}}Department of Electrical and Computer Engineering, Seoul National University \\
\textsuperscript{\textnormal{2}}Interdisciplinary Program in Artificial Intelligence, Seoul National University \\
\textsuperscript{\textnormal{3}}AIIS, ASRI, INMC, and ISRC, Seoul National University 
\textsuperscript{\textnormal{4}}DGIST
\\ \\ 
\texttt{\{tkdrnjs0621, 0k9d0h1, sryoon\}@snu.ac.kr, jmok@dgist.ac.kr}
}

\begin{document}
\maketitle
\begin{abstract}
The conventional Retrieval-Augmented Generation (RAG) paradigm of injecting raw retrieved texts into the Large Language Model (LLM)'s context often results in suboptimal integration of retrieved information.
This paper proposes to bridge retrieval results and the LLM's reasoning ability through Verbal Annotations, analytic narratives that explicitly articulate the logical connection between a search query and retrieved contexts.
Our empirical investigation reveals the potential of Verbal Annotations to substantially enhance the LLM's ability to generate accurate, contextually-grounded responses.
Motivated by this finding, we introduce Verbal-R3, a novel agentic RAG framework that consists of a Generator and a Verbal Reranker.
The Generator performs iterative retrieval and reasoning, while the Verbal Reranker returns relevance scores and Verbal Annotations to guide the reasoning and answering process of the Generator.
The inference process of Verbal-R3 is further refined through relevance-guided test-time scaling, which efficiently allocates test-time compute for effective trajectory expansion.
Verbal-R3 achieves state-of-the-art performance on complex Question Answering benchmarks, validating the effectiveness of the proposed framework.\footnote{\url{https://github.com/0k9d0h1/VerbalR3}}
\end{abstract}

\section{Introduction}

The Large Language Models' (LLMs)~\cite{GPT4,Qwen3} tendency to hallucinate and the inevitable constraints of fixed training cut-offs undermine the reliability and trustworthiness of their outputs~\cite{hallucinationsurvey}.
To mitigate such issues, Retrieval-Augmented Generation (RAG) has naturally emerged as a promising framework~\cite{lewis2020retrieval}.
By enabling an explicit integration of additional or up-to-date knowledge from external databases or sources into the LLMs' generation process, RAG extends the applicability of LLMs to knowledge-intensive tasks. 

The most na\"ive approach to RAG involves directly injecting raw retrieved texts into the LLMs' context~\cite{gao2023retrieval}.
However, this prevailing paradigm in RAG remains suboptimal due to exposure bias, a mismatch between the pre-training data distribution and retrieved contexts, and the frequent inclusion of irrelevant, distracting information.
Overcoming this technical bottleneck necessitates an intermediate module capable of transforming raw retrieval results \cite{Search-o1}.

\begin{figure*}[t]
  \centering
  \includegraphics[width=\linewidth]{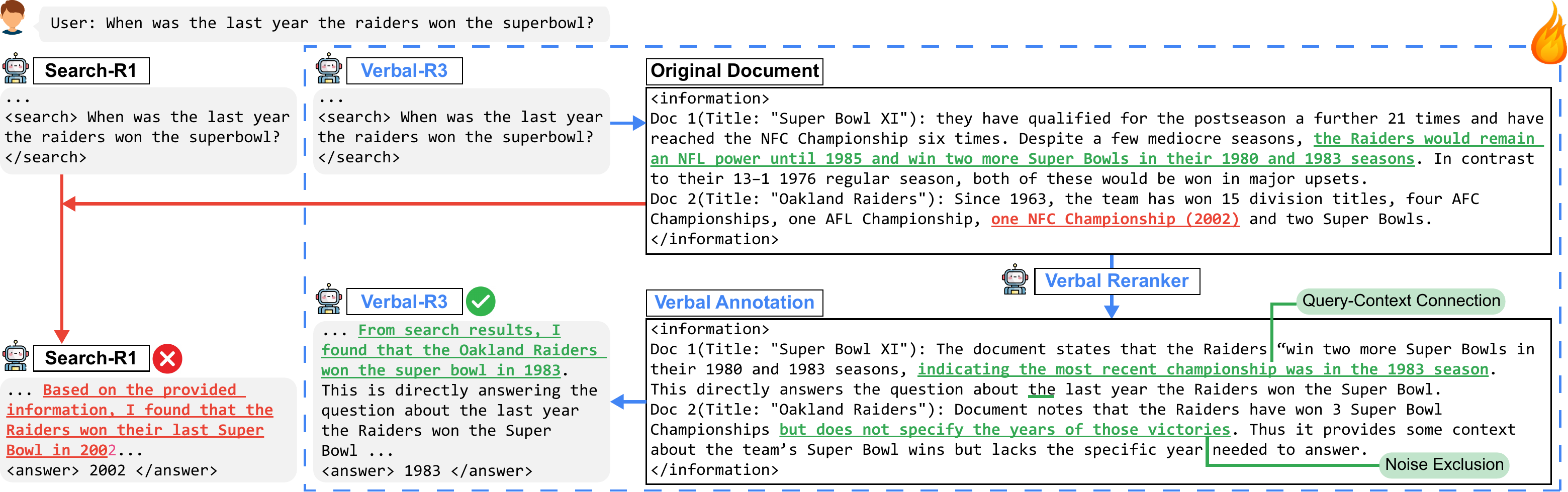}\\[-0.3em]
  \caption{\label{fig:vanilla_vs_ours} Comparison of Search-R1~\textit{vs.} Verbal-R3 inference procedures. The original document contains a large amount of extraneous information, which makes Search-R1 hallucinate and produce an incorrect answer. In contrast, Verbal-R3 successfully exploits Verbal Annotations to distinguish relevant information, allowing the Generator to avoid hallucinations and generate a correct answer.}
\end{figure*}

To bridge this gap, we propose to adopt~\textit{Verbal Annotations}, an analytic narrative that verbalizes the logical connection between the query and the retrieved contexts.
As shown in~\figurename~\ref{fig:vanilla_vs_ours}, Verbal Annotations go beyond merely paraphrasing or summarizing a retrieved document; they state the specific alignment between the query and the document, providing a concrete logical bridge that the LLM can work with.
To elucidate the significance of Verbal Annotations, we conduct an experimental investigation into how various methods of formatting and feeding retrieval results influence the accuracy of the final answer generated by an LLM. 
Interestingly, simply paraphrasing retrieved texts to match the stylistic patterns preferred by an LLM struggles to encourage an effective integration of retrieval results.
On the contrary, Verbal Annotations enhance the correctness of the final answer by stating the relevance of the retrieved information and removing irrelevant noise.

Inspired by the \xx{effectiveness} of Verbal Annotations, this work introduces the concept of a~\textit{Verbal Reranker}: an auxiliary LLM capable of evaluating the relevance of retrieved documents to the query and transforming them into a more digestible form,~\textit{i.e.,} Verbal Annotations.
Our proposed two-part RAG system, dubbed Verbal-R3 (Retrieve-Rerank-Reason), is comprised of a Generator and a Verbal Reranker LLMs.
\xx{The Generator identifies information required to answer the user's question and produces a retrieval query accordingly. 
The Verbal Reranker then scores and reorders retrieved documents, while additionally rewriting each document into a verbalized form that states its relevance. 
This process is repeated until the Generator determines that sufficient information has been obtained to produce the final answer.}

The training pipeline of Verbal-R3 consists of Verbal Reranker distillation and Generator alignment.
Because relying on a proprietary or large reasoning model,~\textit{e.g.,} GPT-OSS-120B \cite{openai2025gptoss120bgptoss20bmodel}, incurs a huge computational cost, we collect query-document-Verbal Annotation triplets from such a model and distill them to smaller models with 1.5B or 3B parameters through supervised fine-tuning.
Afterwards, the Generator is aligned with the trained lightweight Verbal Reranker via Group Relative Policy Optimization (GRPO) \cite{deepseek-math}.
Additionally, at inference time, we newly incorporate relevance-guided test-time scaling that leverages the Verbal Reranker as a query evaluator for the Generator, enabling a computationally efficient yet effective trajectory expansion.

Extensive experiments demonstrate that our lightweight Verbal Reranker more effectively improves RAG performance than existing reranking solutions. 
Moreover, our Verbal-R3 framework surpasses previous state-of-the-art results across a diverse range of Question Answering benchmarks. 
Our contributions are largely three-fold:
\begin{itemize} 
\item{We reveal that Verbal Annotations, structured analytic narratives that map the query-context relationship, are essential for enhancing an LLM's ability to produce accurate responses in knowledge-intensive tasks.}
\item{Building on the effectiveness of Verbal Annotations, we propose Verbal-R3, a novel two-agent RAG framework that interleaves a Generator for iterative retrieval and reasoning and a Verbal Reranker that generates Verbal Annotations to guide the Generator's reasoning.}
\item{We introduce relevance-guided test-time scaling to Verbal-R3 that dynamically assigns test-time computation to the most promising reasoning paths and queries.}
\end{itemize}

\begin{table*}[t]
    \centering
    \resizebox{\textwidth}{!}{
        \begin{tabular}{l|ccccccc|c}
        \toprule
        \textbf{Method} & \textbf{2Wiki} & \textbf{Bamboogle} & \textbf{HotpotQA} & \textbf{MuSiQue} & \textbf{NQ} & \textbf{PopQA} & \textbf{TriviaQA} & \textbf{Average} \\
        \midrule
        \textbf{Search-R1 3B} 
        & 38.06 & 32.75 & 37.50 & 14.69 & 43.49 & 43.98 & 60.75 & 38.75 \\
        
        \textbf{Search-R1 3B + Paraphrase} 
        & 37.38 & 32.05 & 35.42 & 13.94 & 41.94 & 43.11 & 60.59 & 37.78 \\
        
        \textbf{Search-R1 3B + Verbal Annotation} 
        & \textbf{38.65} & \textbf{40.40} & \textbf{39.58} & \textbf{16.01} & \textbf{45.71} & \textbf{45.83} & \textbf{67.28} & \textbf{41.92} \\
        \bottomrule
        \end{tabular}
    }
    \vspace{-0.3em}
    \caption{Analysis of how different context integration methods affect the Search-R1 performance on QA benchmarks.}
    \vspace{-1em}
    \label{table:search_r1_ablation}
\end{table*}

\section{Related Works}
\subsection{Retrieval Augmented Generation}
Retrieval-Augmented Generation (RAG) was introduced to mitigate hallucination, overcome knowledge cutoff limitations, and improve robustness for domain-specific knowledge \citep{lewis2020retrieval, gao2023retrieval}.
Various approaches have been proposed to improve RAG performance, including designing effective retrievers \cite{izacard2021unsupervised, wang2022text, zhang2025qwen3} and enhancing their utility through query and document rewriting techniques \citep{ma2023query, gao2023precise, Search-o1}.

However, a single retrieval step is often insufficient for complex queries requiring multi-hop reasoning. 
To address this limitation, agentic approaches that treat retrieval as a tool have emerged. 
These methods have been explored through both test-time approaches \citep{ircot, iter-retgen,jiang-etal-2023-active, press2023measuring} and training-based methods \citep{asai2024selfrag, yu2024autorag}. 
To address the scarcity of ground-truth reasoning trajectories, recent works have integrated reinforcement learning to optimize search-based reasoning models \citep{searchr1, zheng-etal-2025-deepresearcher}.
While Search-o1~\cite{Search-o1} explores an evaluation of retrieved documents within agentic retrieval, our Verbal Reranker distinguishes itself in that its outputs are structurally designed to capture logical query–context relationships.

\subsection{Rerankers}
Reranking serves as a critical refinement step in retrieval pipelines, where an initial set of candidate documents retrieved by a first-stage system is reordered to prioritize the most relevant items by modeling the joint relation between the query and each document \cite{gao2023retrieval}.

The advent of transformer-based architectures \citep{vaswani2017attention, devlin-etal-2019-bert} marked a significant shift in reranking capabilities. Cross-encoders and late-interaction models \citep{nogueira2019passage, khattab2020colbert} leveraged BERT-based encoders to produce relevance scores, while sequence-to-sequence models using T5 \citep{T5} as their backbone reformulated reranking as a generation task \citep{nogueira2019document, nogueira2020document,zhuang2023rankt5, yoon2024listt5}.
Recent work has explored leveraging large language models for reranking through both zero-shot prompting \citep{sun2023chatgpt, ma2023zero} and supervised fine-tuning \citep{pradeep2023rankvicuna, pradeep2023rankzephyr}. Our work adopts a pointwise Verbal Reranker that generates natural language explanations alongside relevance assessments, thereby providing functionality beyond conventional rerankers.

\section{Motivation}
\label{sec:motivation}

\paragraph{Verbal Annotation.} 
The suboptimal performance of RAG systems is commonly attributed to exposure bias, the distributional shift between an LLM's pre-training data and raw retrieved contexts it encounters at inference time~\citep{ranzato2015sequence, arora-etal-2022-exposure}. 
This chasm between retrieved information and the LLM's reasoning capabilities is further deepened by the presence of tangential noise within raw retrieval results. 
Building on this observation, we posit that~\textit{Verbal Annotations}, analytic statements that explicitly articulate the relevance of the retrieved contexts, can serve as a crucial bridge to improve the fidelity of the final answer. 
Unlike standard context injection, a Verbal Annotation does not merely present additional text; it actively interprets the retrieved context to distinguish relevant evidence from irrelevant noise, articulates precisely why specific segments support the query, and justifies the exclusion of non-pertinent parts.
In doing so, a Verbal Annotation aligns the external data with the internalized logical requirements of the LLM. 

\paragraph{Significance of Verbal Annotations.} 
 To assess the impact of Verbal Annotations, we conduct a comparative analysis using Search-R1~\citep{searchr1} as the backbone LLM, evaluating three distinct methods of feeding retrieved contexts on Question Answering (QA) benchmarks:\\
$\cdot$~\textbf{Na\"ive Context Integration:} Directly injecting original retrieved passages into the LLM's input context with the prompt in~\figurename~\ref{fig:p_naive_rag}. \\
$\cdot$~\textbf{Paraphrased Context:} Rewriting passages using an external module in the stylistic patterns preferred by LLMs, improving their readability without a logical commentary. \\
$\cdot$~\textbf{Verbal Annotations:} Incorporating an analytic narrative that verbalizes the logical connection between the query and the documents.



In our exploratory experiments, both the Paraphrased Context and Verbal Annotations are obtained by prompting the GPT-OSS-120B model \citep{openai2025gptoss120bgptoss20bmodel} using the templates in Figures~\ref{fig:p_paraphrase} and~\ref{fig:p_verbal_reranker}. The prompt template used to generate Verbal Annotations was deliberately refined to produce higher-quality annotations, and the refinement process is further detailed in Section ~\ref{sec:reranker_training}.
To complement Verbal Annotations, GPT-OSS-120B additionally assigns a numerical relevance score to each document through a point-wise reranking approach, which assesses each document separately.

Our results in~\tablename~\ref{table:search_r1_ablation} demonstrate that Paraphrased Context underperforms Na\"ive Context Integration, whereas employing Verbal Annotations significantly outperforms both baselines.
This result suggests that simply smoothing the text distribution through paraphrasing may inadvertently filter out crucial details or introduce unintended hallucinations.
In contrast, Verbal Annotations act as a rigorous filter that explicitly elaborates the relationship between the query and the document. 
By analyzing which parts of the document are relevant, they prevent the LLM from being distracted by noise, serving as a cognitive bridge rather than a superficial stylistic transfer.

\begin{table}
\centering
\footnotesize
\resizebox{1\linewidth}{!}{
\begin{NiceTabular}{l|cccc|cc}
\toprule

Model & EM & F1 & RA-R & RA-L & $\mathtt{CUE}$-$\mathtt{R}$ & $\mathtt{CUE}$-$\mathtt{L}$ \\
\midrule

\textit{w/o} VA
& 38.75 & 46.31 & 61.79 & \textbf{48.26} & 64.77 & 69.04 \\

\textit{w/} VA
& \textbf{41.92} & \textbf{49.86} & \textbf{62.84} & 46.37 & \textbf{70.76} & \textbf{74.10} \\
\bottomrule

\end{NiceTabular}
}
\vspace{-0.3em}
\caption{Analysis of Search-R1 with and without Verbal Annotation (VA). RA denotes Retrieval Accuracy.}
\vspace{-1em}
\label{table:search_r1_verbal_annotation}
\end{table}

\paragraph{Analysis of Verbal Annotations.}
To evidence that Verbal Annotations improve the correctness of the final answer by facilitating an effective incorporation of retrieval results, we explore the effect of Verbal Annotations through a new metric, Context Utilization Efficacy ($\mathtt{CUE}$).
$\mathtt{CUE}$ is computed by the conditional probability of the final answer being correct given that the retrieval result is accurate:
\begin{equation*}
    \mathtt{CUE} = P(\text{Answer Correct} \mid \text{Retrieval Accurate}).
\end{equation*}
The Retrieval Accuracy (RA) can be measured via either a rule-based heuristic (RA-R), based on whether the answer is included in the context, or an LLM-as-Judge evaluation (RA-L), where an external large LLM assesses whether crucial evidence for answering the question is present. 
We use GPT-4.1 \cite{openai2025gpt41} as the Judge model with the prompt in~\figurename~\ref{fig:p_cue_l}.
Given RA-R and RA-L, we implement two variants of $\mathtt{CUE}$: $\mathtt{CUE}$-$\mathtt{R}$ and $\mathtt{CUE}$-$\mathtt{L}$.

\tablename~\ref{table:search_r1_verbal_annotation} shows that Verbal Annotations substantially increase both $\mathtt{CUE}$-$\mathtt{R}$ and $\mathtt{CUE}$-$\mathtt{L}$.
We note that Verbal Annotations improve $\mathtt{CUE}$-$\mathtt{R}$ and $\mathtt{CUE}$-$\mathtt{L}$ more than they do answering (Exact Match \& F1) and retrieval accuracies (RA).  
This particular increase in $\mathtt{CUE}$ from Verbal Annotations confirms that the structured cognitive scaffold provided by them (\textit{e.g.,} an explicit mention of context relevance) effectively guides the reasoning process of LLMs.

\section{Methodology}
Our results from the previous section strongly motivate the need for an external~\textit{Verbal Reranker} that not only evaluates the relevance of each retrieved context but also refines it into Verbal Annotations.
The mechanism of Verbal Annotations echoes the logic of neural rerankers, which evaluate query-document compatibility~\cite{khattab2020colbert}.
We now present Verbal-R3, a two-part RAG framework that consists of a Generator, an LLM that performs iterative retrieval and reasoning, and a compact Verbal Reranker.

First, we train a lightweight Verbal Reranker by distilling query-document-Verbal Annotation triplets obtained from a larger, more performant reasoning model.
Then, we align the Generator LLM to maximally utilize Verbal Annotations from the Verbal Reranker, and finally, we leverage a relevance-guided test-time scaling approach that strategically allocates computational resources to achieve performance gains and efficiency.

\begin{figure*}[t]
  \centering
  \includegraphics[width=\linewidth]{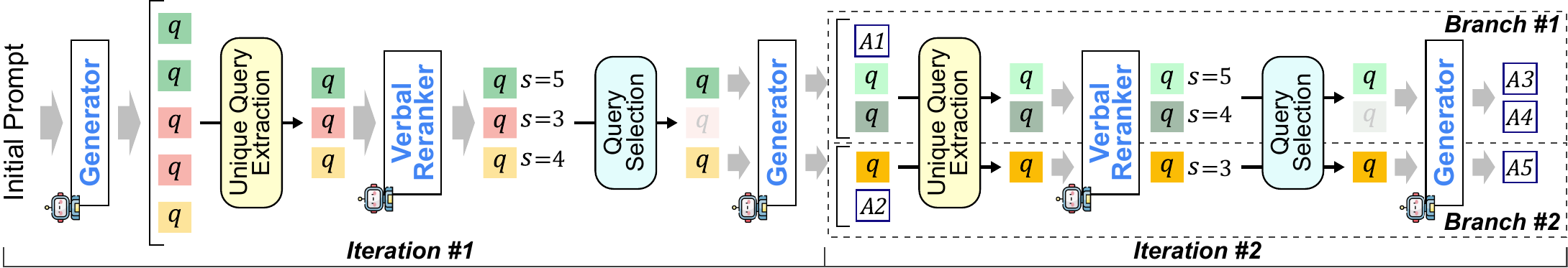}\\[-0.3em]
  \caption{\label{fig:our_testtime_scaling}  An example of relevance-guided test-time scaling with two iterations and two branches. Different background colors for queries ($q$) represent distinct queries and $A1$, \dots, $A5$ denote the generated answers.}
  \vspace{-1em}
\end{figure*}

\subsection{Training Lightweight Verbal Reranker}
\label{sec:reranker_training}
Calling proprietary models or large open-weight models as a Verbal Reranker every time a Verbal Annotation is necessary requires an enormous amount of compute, rendering them impractical for deployment.
This critical limitation accentuates the need for an lightweight alternative to enable efficient decision-making within the iterative retrieval loop.
We utilize GPT-OSS-120B as the teacher model to synthesize Verbal Annotations, which are then distilled into lightweight student models with 1.5B or 3B trainable parameters suitable for high-throughput inference.

The quality of distilled annotations is highly sensitive to the teacher model's instruction design. 
An extensive prompt search is performed to identify the optimal prompting strategy, evaluating candidate prompts based on their ability to elicit discriminative and critical feedback rather than generic summaries. 
The best-performing prompt in~\figurename~\ref{fig:p_verbal_reranker} explicitly instructs the teacher model to assign a relevance score and fill logical gaps in retrieved documents. 

Using this optimized configuration, we generate a large-scale corpus of synthetic query-document-Verbal Annotation triplets. 
Each Verbal Annotation includes both a textual critique and a scalar relevance score on a 1--5 scale, which is used for reranking documents. 
The quality of the resulting data is further validated by Gemini 2.5 Pro \cite{comanici2025gemini25pushingfrontier}, a near-human-level proprietary LLM from a different model family, tasked with assessing the reasonability of each generated Verbal Annotation, yielding a 98.5\% approval rate.

A Verbal Reranker is trained with a supervised fine-tuning (SFT) objective over synthetic Verbal Annotations from the teacher model. 
The input is $
x^\mathrm{VR}=(i_{\mathrm{VR}},q,d)$, where $i_{\mathrm{VR}}$, $q$, $d$, denote the assessing prompt, a query, and a document, respectively.
The output is $y^\mathrm{VR}=(v,s)$, where $v$ and $s$ denote a Verbal Annotation and a relevance score ($s\in\{1,2,3,4,5\}$).
Given a dataset $\mathcal{D}=\{(x^\mathrm{VR},y^\mathrm{VR})\}$, the training objective is expressed as:
\begin{equation*}
\begin{aligned}
\mathcal{L}_{\mathrm{SFT}}(\theta)
&= -\mathbb{E}_{(x^\mathrm{VR},y^\mathrm{VR})\sim\mathcal{D}}
\\
&\qquad \Bigg[ \sum_{t=1}^{T}
\log p_\theta\!\left(
y^\mathrm{VR}_t
\mid x^\mathrm{VR}, y^\mathrm{VR}_{<t}
\right)
\Bigg],
\end{aligned}
\end{equation*}
where $p_\theta$ denotes the student Verbal Reranker parameterized by $\theta$, and $T$ is the output sequence length.
This results in a compact module capable of emulating the nuanced judgment of a 120B model while maintaining the low latency required for iterative retrieval.

\subsection{Aligning Generator with Verbal Reranker via Reinforcement Learning}

\noindent{\textbf{Preliminary.}} 
Following the training of the Verbal Reranker, we optimize the Generator to maximally utilize Verbal Annotations from the Verbal Reranker.
Following the training protocol in Search-R1~\cite{searchr1}, the Generator policy $\pi_\phi$ is trained via Group Relative Policy Optimization GRPO~\cite{deepseek-math} to maximize the objective below:
{
\setlength{\abovedisplayskip}{5pt}
\setlength{\belowdisplayskip}{5pt}
\setlength{\abovedisplayshortskip}{0pt}
\setlength{\belowdisplayshortskip}{0pt}
\begin{equation*}
\label{eq:grpo}
\footnotesize
\begin{aligned}
    \mathcal{J}_{\mathrm{GRPO}}(\phi) &= \mathbb{E}_{\substack{x^{\mathrm{Gen}} \sim \mathcal{D} \\ \{y^{\mathrm{Gen}}_i\} \sim \pi_{\mathrm{old}}}} \biggl[ \frac{1}{G} \sum_{i=1}^G \frac{1}{\sum_{t=1}^{|y^{\mathrm{Gen}}_i|} I(y^{\mathrm{Gen}}_{i,t})} \\[-1.0ex]
    \sum_{t:I(y^{\mathrm{Gen}}_{i,t})=1}^{|y^{\mathrm{Gen}}_i|} & \quad \min \left(r_{i,t} \hat{A}_{i}, \mathrm{clip} (r_{i,t}, 1-\epsilon, 1+\epsilon ) \hat{A}_{i} \right) \\[-1.0ex]
    & \quad - \beta \mathbb{D}_{\mathrm{KL}} [\pi_\phi || \pi_{\mathrm{ref}}] \biggr],
\end{aligned}
\end{equation*}
}
where $I(y^{\mathrm{Gen}}_{i,t})$ is the indicator function that yields 1 only for tokens generated by the policy model being trained and 0 otherwise (\textit{e.g.,} reranker responses). 
$r_{i,t}$ denotes the importance ratio, defined as $\frac{\pi_\phi(y^{\mathrm{Gen}}_{i,t} | x^{\mathrm{Gen}}, y^{\mathrm{Gen}}_{i,<t})}{\pi_{\text{old}}(y^{\mathrm{Gen}}_{i,t} | x^{\mathrm{Gen}}, y^{\mathrm{Gen}}_{i,<t})}$, and $\hat{A}_{i}$ represents the group-relative advantage, which is computed by normalizing the sequence-level reward $R_i$ within each group: $\hat{A}_{i} = \frac{R_i - \text{mean}(\{R_k\}_{k=1}^G)}{\text{std}(\{R_k\}_{k=1}^G)}$.

\begin{table*}[!t]
    \centering
    \resizebox{\textwidth}{!}{
        \begin{tabular}{lcccccccccccccc|cc}
        \toprule
        \multirow{2}{*}{\textbf{Method}} 
        & \multicolumn{2}{c}{\textbf{2Wiki}} 
        & \multicolumn{2}{c}{\textbf{Bamboogle}} 
        & \multicolumn{2}{c}{\textbf{HotpotQA}} 
        & \multicolumn{2}{c}{\textbf{MuSiQue}} 
        & \multicolumn{2}{c}{\textbf{NQ}} 
        & \multicolumn{2}{c}{\textbf{PopQA}} 
        & \multicolumn{2}{c|}{\textbf{TriviaQA}} 
        & \multicolumn{2}{c}{\textbf{Average}} \\
        \cmidrule(lr){2-3} \cmidrule(lr){4-5} \cmidrule(lr){6-7} \cmidrule(lr){8-9}
        \cmidrule(lr){10-11} \cmidrule(lr){12-13} \cmidrule(lr){14-15} \cmidrule(lr){16-17}
        & \textbf{EM} & \textbf{F1}
        & \textbf{EM} & \textbf{F1}
        & \textbf{EM} & \textbf{F1}
        & \textbf{EM} & \textbf{F1}
        & \textbf{EM} & \textbf{F1}
        & \textbf{EM} & \textbf{F1}
        & \textbf{EM} & \textbf{F1}
        & \textbf{EM} & \textbf{F1} \\
        \midrule
        \textbf{E5 (Baseline)} 
        & 13.54 & 18.83 & 08.90 & 16.93 & 23.55 & 32.75 & 04.10 & 08.27 & 33.49 & 44.51 & 37.04 & 43.98 & 53.16 & 62.50 & 24.83 & 32.54 \\
        
        \textbf{MonoT5 3B} 
        & 17.13 & 22.57 & 12.65 & 18.39 & 28.72 & 39.38 & 06.87 & 12.99 & 33.57 & 44.62 & 37.95 & 44.90 & 56.94 & 66.78 & 27.69 & 35.66 \\
        
        \textbf{RankLLama 7B} 
        & 17.54 & 23.39 & 12.60 & 19.32 & \textbf{28.44} & \textbf{39.05} & 06.74 & 12.27 & 34.60 & 46.50 & 39.55 & 46.71 & 57.69 & 67.56 & 28.17 & 36.40 \\
        
        \textbf{Rank1 7B} 
        & 16.86 & 22.16 & 15.75 & 23.60 & 27.70 & 38.07 & \textbf{06.95} & 12.68 & \textbf{35.82} & \textbf{47.65} & 39.71 & 46.73 & 59.38 & 69.46 & 28.88 & 37.19 \\

        \midrule
        
        \textbf{Ours 1.5B} 
        & 20.20 & 24.66 & 15.25 & 22.86 & 24.13 & 33.93 & 05.75 & 11.42 & 33.91 & 46.02 & 39.38 & 46.84 & 58.00 & 67.45 & 28.09 & 36.17 \\
        
        \textbf{Ours 3B} 
        & \textbf{21.74} & \textbf{26.31} & \textbf{20.35} & \textbf{29.64} & 26.47 & 36.99 & 06.91 & \textbf{12.97} & 34.90 & 47.51 & \textbf{40.77} & \textbf{48.41} & \textbf{60.16} & \textbf{70.02} & \textbf{30.19} & \textbf{38.84} \\
        \bottomrule
        \end{tabular}
    }
    \caption{Evaluation results of single-turn retrieval systems augmented with various reranker models on Question-Answering benchmarks. Qwen-2.5-3B-Instruct is used as the model for answer generation.}
    \label{table:main_reranker_rag}
\end{table*}

\noindent{\textbf{Trajectory Construction.}} The output sequence $y^{\mathrm{Gen}}$ is constructed iteratively through multi-turn interactions between the Generator $\pi_\theta$ and the Verbal Reranker.
When the Generator produces a search query $q$ after reasoning about the user's question, a retriever first fetches a set of $n$ documents $\mathcal{D} = \{d_1, d_2, \dots, d_n\}$.
For each document $d_i$, the Verbal Reranker evaluates its relevance to $q$ and produces a Verbal Annotation $v_i$ and an integer relevance score $s_i \in \{1, 2, 3, 4, 5\}$:
\vspace{-0.3em}
\begin{equation*}
(v_i, s_i) \sim p_{\mathrm{\theta}}(\,\cdot \,| \,i_{\mathrm{VR}}, q, d_i).
\end{equation*}
To further distinguish between documents with identical scores, we extract the logit value $l_i$ corresponding to the predicted score token $s_i$ from the Verbal Reranker's output distribution.

From the candidate pool, we select the top-$k$ Verbal Reranker responses based on the assigned scores $s_i$. In the event of a tie, the response with the higher generation logit ${l_i}$ is prioritized.
The selected responses are then appended to the sequence $y$ in the format: "<information>\dots[Doc $i$] $v_i$ (Relevance score: $s_i$)\dots</information>".
This iterative process of reasoning, query generation, retrieval, and reranked context augmentation continues until a final answer is successfully parsed from the Generator's output.

\noindent{\textbf{Reward Design.}} To simultaneously incentivize factual accuracy and structural consistency, we define a hierarchical reward function similar to \cite{searchr1ver2}. 
The reward comprises the outcome reward ($R_{\mathrm{outcome}}$), the format reward ($R_{\mathrm{format}}$), and the answer reward ($R_{\mathrm{answer}}$).
The total reward $R$ is assigned based on the correctness of the final answer and its adherence to the prescribed format:
\begin{equation*}
\small
R = 
\begin{cases} 
R_{\mathrm{outcome}} & \text{\scriptsize Correct Answer \& Correct Format} \\
R_{\mathrm{outcome}} - R_{\mathrm{format}} & \text{\scriptsize Correct Answer \& Incorrect Format} \\
R_{\mathrm{format}} & \text{\scriptsize Incorrect Answer \& Correct Format} \\
R_{\mathrm{answer}} & \text{\scriptsize Incorrect Format \& Answer Parsed} \\
0.0 & \text{\scriptsize Otherwise}
\end{cases}
\end{equation*}
Detailed parsing methods and formats are provided in Section~\ref{sec:appendix Parsing Methods}.

\subsection{Relevance-Guided Test-Time Scaling}
We propose to leverage the relevance score assigned by the Verbal Reranker as a proxy for evaluating the quality of queries produced by the Generator at inference time.
This approach is grounded in the premise that a high relevance score validates the Generator’s information-seeking strategy by indicating that the query successfully guided the retriever to pertinent documents.
Based on this intuition, we introduce an efficient relevance-guided test-time scaling method that selectively explores effective search trajectories.

We define each iteration $t$ as a set of branches, where a single iteration encompasses a complete cycle of reasoning, query generation, retrieval, and reranking.
Within this framework, each branch represents an independent reasoning trajectory that maintains its own accumulated historical context and serves as the fundamental unit for the query selection and path expansion processes.

\noindent{\textbf{Scoring and Selection.}}
After the generation, parsed answers are added into the answer candidate set $\mathcal{F}$, while generated queries are utilized for the selection process described below.
To selectively explore promising Generator trajectories, our method prioritizes the extraction of effective search directions, unlike na\"ive majority voting~\cite{selfconsistency} which treats all trajectories equally.


To optimize computational efficiency, retrieval and reranking are performed only for the unique query set.
For each unique query, we obtain a set of relevance scores $\mathcal{S}_q=\{s_1,s_2,\dots,s_n\}$ and the logit $l_q$ of $s_{\max, q}$ via the Verbal Reranker, defining $s_{\max, q} = \max(\mathcal{S}_q)$.

In the sampling step, we consider the unique query set for each branch.
A query $q$ and corresponding Verbal Annotation is selected for the next iteration with a probability $P(q)$ that reflects its relative strength within the branch:
{
\setlength{\abovedisplayskip}{6pt}
\setlength{\belowdisplayskip}{6pt}
\setlength{\abovedisplayshortskip}{0pt}
\setlength{\belowdisplayshortskip}{0pt}
$$P(q) = \left( \frac{s_{\max, q}}{s_{\mathrm{best}}} \right)^\alpha, $$
}
where $s_{\mathrm{best}}$ is the highest score achieved within a branch, and $\alpha$ is a scaling hyperparameter.
If a query is selected, a new branch for the subsequent iteration is initialized with an augmented context, constructed by appending the reasoning, the selected query, and its corresponding Verbal Annotation to the historical context of the current branch.
This ensures that generation contexts are predominantly constructed along promising reasoning paths without sacrificing diversity across reasoning trajectories.

Following query selection, the total number of generations across all active branches is regulated to match the remaining budget, $N - |\mathcal{F}|$, where $N$ is a total trajectory budget. 
We distribute these sample slots to ensure each branch receives as uniform a base allocation as possible.
Any remaining slots from the division are sequentially assigned to branches in descending order of their relevance scores $s_{max,q}$ and logits $l_{q}$, both of which were previously stored during the reranking.

\noindent{\textbf{Consolidation.}}
This process iterates until $|\mathcal{F}| = N$ or maximum iteration $T$ is reached.
Finally, majority voting is performed over the answers collected in $\mathcal{F}$ to determine the final output $A^*$.
We provide the detailed workflow of relevance-guided test-time scaling in Algorithm~\ref{alg:tts}, as well as an illustrative example in~\figurename~\ref{fig:our_testtime_scaling}.

\begin{table*}[t]
    \centering
    \resizebox{\textwidth}{!}{\leavevmode
        \begin{tabular}{lcccccccccccccc|cc}
        \toprule
        \multirow{2}{*}{\textbf{Method}} & \multicolumn{2}{c}{\textbf{2wiki}} & \multicolumn{2}{c}{\textbf{Bamboogle}} & \multicolumn{2}{c}{\textbf{HotpotQA}} & \multicolumn{2}{c}{\textbf{MuSiQue}} & \multicolumn{2}{c}{\textbf{NQ}} & \multicolumn{2}{c}{\textbf{PopQA}} & \multicolumn{2}{c|}{\textbf{TriviaQA}} & \multicolumn{2}{c}{\textbf{Average}} \\ \cmidrule(lr){2-3} \cmidrule(lr){4-5} \cmidrule(lr){6-7} \cmidrule(lr){8-9} \cmidrule(lr){10-11} \cmidrule(lr){12-13} \cmidrule(lr){14-15} \cmidrule(lr){16-17}
        & \textbf{EM} & \textbf{F1} & \textbf{EM} & \textbf{F1} & \textbf{EM} & \textbf{F1} & \textbf{EM} & \textbf{F1} & \textbf{EM} & \textbf{F1} & \textbf{EM} & \textbf{F1} & \textbf{EM} & \textbf{F1} & \textbf{EM} & \textbf{F1} \\ \midrule
        \textit{\textbf{Qwen2.5-3B}} \\
        Direct Answer & 18.47 & 22.89 & 3.20 & 7.87 & 12.23 & 19.08 & 1.57 & 6.78 & 9.14 & 15.21 & 7.02 & 11.01 & 22.73 & 30.51 & 10.62 & 16.19 \\
        RAG & 13.54 & 18.83 & 8.90 & 16.93 & 23.55 & 32.75 & 4.10 & 8.27 & 33.49 & 44.51 & 37.04 & 43.98 & 53.16 & 62.50 & 24.83 & 32.54 \\ 
        IRCoT & 17.37 & 24.89 & 16.90 & 25.10 & 21.82 & 30.60 & 5.09 & 9.95 & 23.13 & 33.39 & 32.23 & 39.33 & 40.18 & 49.77 & 22.26 & 30.49 \\ 
        ITER-RETGEN & 23.83 & 29.41 & 12.55 & 21.51 & 28.01 & 37.21 & 5.96 & 11.21 & 36.81 & 46.55 & 41.61 & 47.02 & 55.86 & 64.80 & 29.23 & 36.82 \\ 
        Search-R1 & 38.06 & 44.10 & 32.75 & 42.73 & 37.50 & 47.83 & 14.69 & 22.36 & 43.49 & 51.68 & 43.98 & 47.94 & 60.75 & 67.52 & 38.75 & 46.31 \\ 
        Verbal-R3 (Ours) & \textbf{43.84} & \textbf{51.00} & \textbf{49.25} & \textbf{61.34} & \textbf{44.47} & \textbf{56.95} & \textbf{20.89} & \textbf{29.99} & \textbf{46.01} & \textbf{56.31} & \textbf{47.59} & \textbf{52.85} & \textbf{65.50} & \textbf{74.15} & \textbf{45.36} & \textbf{54.66} \\ \midrule
        \textit{\textbf{Qwen2.5-7B}} \\
        Direct Answer & 23.12 & 28.47 & 10.25 & 16.89 & 17.14 & 25.07 & 3.60 & 11.69 & 12.58 & 21.63 & 14.54 & 18.89 & 38.02 & 45.35 & 17.04 & 24.00 \\ 
        RAG & 20.46 & 25.76 & 14.15 & 23.29 & 27.64 & 36.62 & 5.13 & 9.39 & 31.47 & 42.31 & 37.79 & 45.16 & 55.75 & 64.52 & 27.48 & 35.92 \\ 
        IRCoT & 14.02 & 21.84 & 18.30 & 28.67 & 23.89 & 33.98 & 5.05 & 10.22 & 23.71 & 37.26 & 32.51 & 40.62 & 49.85 & 60.75 & 23.90 & 33.33 \\ 
        ITER-RETGEN & 24.99 & 31.74 & 20.65 & 30.75 & 31.72 & 42.15 & 6.99 & 14.37 & 36.37 & 47.41 & 40.91 & 47.80 & 60.02 & 69.10 & 31.66 & 40.47 \\ 
        Search-R1 & 40.17 & 46.55 & 40.90 & 50.95 & 41.22 & 52.77 & 18.78 & 27.94 & 44.93 & 53.51 & 44.23 & 48.30 & 63.10 & 70.71 & 41.90 & 50.10 \\ 
        Verbal-R3 (Ours) & \textbf{47.19} & \textbf{53.89} & \textbf{53.55} & \textbf{65.25} & \textbf{47.47} & \textbf{60.32} & \textbf{25.16} & \textbf{34.85} & \textbf{48.28} & \textbf{57.87} & \textbf{48.30} & \textbf{52.77} & \textbf{68.13} & \textbf{76.03} & \textbf{48.30} & \textbf{57.28} \\ \bottomrule
        \end{tabular}
    }
    \vspace{-0.3em}
    \caption{Evaluation of RAG systems on single- and multi-hop Question-Answering benchmarks. Comparison with recent agentic methods is included in \ref{sec:appendix Comparison Agentic}.}
    \vspace{-1em}
    \label{table:main_planner}
\end{table*}

\section{Experimental Results and Discussion}

\subsection{Experimental Details}
\noindent{\textbf{Benchmarks.}}
We primarily evaluate models across seven QA benchmarks, categorized into single-hop tasks, namely NQ~\cite{nq}, TriviaQA~\cite{triviaqa}, and PopQA~\cite{popqa}; and multi-hop reasoning tasks, comprising 2WikiMultiHopQA~\cite{2wiki}, Bamboogle~\cite{bamboogle}, MuSiQue~\cite{musique}, and HotpotQA~\cite{hotpotqa}.
For all datasets, performance is measured using Exact Match (EM) and F1 score.
Since the Bamboogle dataset consists of only 125 samples, we report the average performance over 16 sampling trials to ensure the statistical robustness and reliability of our evaluation results. 
To separately assess the reranking ability, the BEIR benchmark~\cite{thakur2021beir} is used, with nDCG@10 as the main evaluation metric for reranking performance.

\noindent{\textbf{Baselines.}}
To evaluate the performance of our Verbal Reranker, we compare it against existing reranker models: MonoT5~\citep{nogueira2019document}, RankLLaMA~\citep{ma2024fine}, and Rank1~\citep{weller2025rank1}.
To evaluate the effectiveness of Verbal-R3 as a standalone RAG system, we compare it against representative baselines in RAG: Direct Answer, Standard RAG, IRCoT~\cite{ircot}, ITER-RETGEN~\cite{iter-retgen}, and Search-R1~\cite{searchr1}.
Implementation details for these baselines are provided in Section~\ref{sec:appendix baselines}. 
Finally, our test-time scaling approach is compared against two variations of na\"ive Majority Voting, whose descriptions are also given in Section~\ref{sec:appendix baselines}.

\noindent{\textbf{Training \& Inference Settings.}}
Following \citet{searchr1}, we use E5 \cite{wang2022text} as our base retriever across all experiments.\\
$\cdot$~\textbf{Verbal Reranker Training}: We distill the outputs of GPT-OSS-120B on the Natural Question(NQ) training set to Qwen2.5-Instruct Model (1.5B \& 3B) \cite{qwen2025qwen25technicalreport}. \\
$\cdot$~\textbf{Generator Training:} As the Generator of Verbal-R3, we employ Qwen2.5-3B and Qwen2.5-7B \cite{qwen2025qwen25technicalreport} as backbones, using NQ and HotpotQA as the training datasets.\\
$\cdot$~\textbf{Verbal-R3 Inference:} The 3B Verbal Reranker is used as the default Verbal Reranker for Verbal-R3, and $n=15$ and $k=3$ are used. Thus, Verbal-R3 3B and 7B denote that 3B and 7B Generators are used, respectively, with the size of the Verbal Reranker fixed as 3B. For test-time scaling methodology, we configure the scaling hyperparameter $\alpha$ to 7.5 and set the total trajectory budget $N$ to 5.

More details on training and inference are provided in Sections~\ref{sec:appendix Training Details} and~\ref{sec:appendix Inference Details}.

\begin{table*}[t]
    \centering
    \resizebox{\textwidth}{!}{
        \begin{tabular}{lcccccccccccccc|cc|c}
        \toprule
        \multirow{2}{*}{\textbf{Method}} & \multicolumn{2}{c}{\textbf{2wiki}} & \multicolumn{2}{c}{\textbf{Bamboogle}} & \multicolumn{2}{c}{\textbf{HotpotQA}} & \multicolumn{2}{c}{\textbf{MuSiQue}} & \multicolumn{2}{c}{\textbf{NQ}} & \multicolumn{2}{c}{\textbf{PopQA}} & \multicolumn{2}{c|}{\textbf{TriviaQA}} & \multicolumn{3}{c}{\textbf{Average}} \\ \cmidrule(lr){2-3} \cmidrule(lr){4-5} \cmidrule(lr){6-7} \cmidrule(lr){8-9} \cmidrule(lr){10-11} \cmidrule(lr){12-13} \cmidrule(lr){14-15} \cmidrule(lr){16-18}
        & \textbf{EM} & \textbf{F1} & \textbf{EM} & \textbf{F1} & \textbf{EM} & \textbf{F1} & \textbf{EM} & \textbf{F1} & \textbf{EM} & \textbf{F1} & \textbf{EM} & \textbf{F1} & \textbf{EM} & \textbf{F1} & \textbf{EM} & \textbf{F1} & \textbf{\# Calls} \\ \midrule
        \textit{\textbf{Verbal-R3 3B}} \\
        No M.V. & 43.84 & 51.00 & 49.25 & 61.34 & 44.47 & 56.95 & 20.89 & 29.99 & 46.01 & 56.31 & 47.59 & 52.85 & 65.50 & 74.15 & 45.36 & 54.66 & 2.27 \\
        Naïve M.V. \\ 
        \quad w/o U.Q.E. & \textbf{46.18} & \textbf{53.10} & \textbf{51.45} & \textbf{62.69} & 46.52 & 58.96 & 23.46 & 32.64 & 47.76 & \textbf{57.76} & \textbf{48.62} & \textbf{53.89} & \textbf{66.83} & \textbf{75.24} & \textbf{47.26} & 56.33 & 11.28 \\ 
        \quad w/ U.Q.E. & 45.17 & 52.15 & 50.40 & 62.48 & 46.06 & 58.45 & 23.42 & 32.32 & 47.06 & 57.33 & 48.54 & 53.69 & 66.20 & 74.74 & 46.69 & 55.88 & 6.52 \\ 
        Relevance-Guide (Ours) & 45.53 & 52.57 & 50.60 & 63.01 & \textbf{46.58} & \textbf{59.29} & \textbf{24.99} & \textbf{33.99} & \textbf{47.95} & \textbf{57.76} & 48.22 & 53.54 & 66.42 & 74.99 & 47.18 & \textbf{56.45} & \textbf{6.18}\\ \midrule
        \textit{\textbf{Verbal-R3 7B}} \\ 
        No M.V. & 47.19 & 53.89 & \textbf{53.55} & 65.25 & 47.47 & 60.32 & 25.16 & 34.85 & 48.28 & 57.87 & 48.30 & 52.77 & 68.13 & 76.03 & 48.30 & 57.28 & 1.82 \\
        Naïve M.V. \\ 
        \quad w/o U.Q.E. & 49.67 & 56.35 & 53.25 & 65.75 & 50.10 & 62.74 & 27.35 & 37.41 & \textbf{50.00} & \textbf{59.52} & \textbf{49.90} & \textbf{54.22} & 69.25 & 77.14 & 49.93 & 59.02 & 9.11 \\ 
        \quad w/ U.Q.E. & 48.95 & 55.71 & 52.45 & 65.20 & 49.44 & 61.91 & 26.77 & 36.79 & 49.31 & 58.91 & 49.11 & 53.53 & 68.60 & 76.42 & 49.23 & 58.35 & 4.63 \\ 
        Relevance-Guide (Ours) & \textbf{49.90} & \textbf{56.66} & 53.45 & \textbf{65.80} & \textbf{50.32} & \textbf{63.22} & \textbf{28.96} & \textbf{38.69} & 49.25 & 58.93 & 49.71 & 54.20 & \textbf{69.27} & \textbf{77.24} & \textbf{50.12} & \textbf{59.25} & \textbf{4.21} \\  \bottomrule
        \end{tabular}
    }
    \vspace{-0.3em}
    \caption{Comparison of different test-time scaling approaches. $N=5$ is used for our relevance-guided approach. M.V. denotes majority voting and U.Q.E. denotes unique query extraction.}
    \label{table:main_test_time_scaling}
\end{table*}

\subsection{Results}

\noindent\textbf{Reranker.} 
We separately assess the effectiveness of our Verbal Reranker by combining it with single-turn RAG frameworks in a plug-and-play manner.
The results, reported in Table~\ref{table:main_reranker_rag} in terms of EM and F1 scores on QA benchmarks, indicate that leveraging our 1.5B-parameter Verbal Reranker achieves performance comparable to larger 3B- and 7B-parameter models, and using our 3B-parameter Verbal Reranker consistently outperforms all baselines.
We additionally compare the stand-alone reranking performance of our Verbal Reranker on the BEIR benchmark. 
The results, along with the discussion, are provided in Section \ref{sec:appendix eval rerank}.

\noindent{\textbf{Verbal-R3.}}
Table~\ref{table:main_planner} presents EM and F1 scores of the Verbal-R3 framework and baselines on seven QA benchmarks.
Verbal-R3 robustly achieves state-of-the-art performance across all evaluated benchmarks, outperforming all baselines, including Search-R1, the most competitive of them.
Notably, Verbal-R3 3B, which utilizes a total of $\sim6$B parameters (3B Generator $+$ 3B Verbal Reranker) not only yields a 17.1\% increase in EM and an 18.0\% improvement in F1 over Search-R1 3B but also surpasses the Search-R1 7B by 8.26\% in EM and 9.10\% in F1.
As for Verbal-R3 7B, it improves EM and F1 of Search-R1 7B by 15.3\% and 14.3\%, respectively.

We also highlight that Verbal-R3 exhibits noticeably larger improvements on multi-hop benchmarks.
Verbal-R3 3B model improves the average F1 score on multi-hop tasks by 26.91\%, which is nearly threefold the 9.67\% gain observed in single-hop tasks.
This trend remains consistent for the 7B model, with a 20.26\% improvement for multi-hop versus 8.20\% for single-hop.
These results indicate that mitigating the noise and redundancy in raw retrieval observations with our Verbal Reranker is particularly crucial in multi-hop scenarios, where the Generator’s context becomes increasingly saturated with a vast amount of external data accumulated through successive search iterations.

Experiments on a train-free variant of Verbal-R3 with a larger backbone model in \ref{sec:appendix scaling} further demonstrate the scalability and cost-efficiency of our method.

\noindent{\textbf{Relevance-Guided Test-time Scaling.}}
Table \ref{table:main_test_time_scaling} shows the results of using our test-time scaling method.
When compared to the na\"ive majority voting without unique query extraction, our relevance-guided method achieves comparable or superior performance while reducing the number of reranker calls by 45.2\% and 53.8\% for the 3B and 7B models, respectively.
Furthermore, our approach outperforms the na\"ive majority voting with unique query extraction, demonstrating a robust reduction in reranker calls alongside improved accuracy.

To quantify the computational efficiency, we empirically measure the average number of generated tokens by the reranker, amounting to approximately 990 tokens per call when $n=15$.
By adopting our prioritized selection strategy, we save approximately 376 and 4,950 tokens per query compared to the naïve baselines with and without unique query extraction, respectively.
These results collectively demonstrate that the relevance-guided test-time scaling effectively prioritizes promising queries and eliminates unnecessary operations, thereby ensuring a highly efficient inference process without compromising, and in many cases enhancing, the overall RAG performance. 

\subsection{Analysis}

\noindent{\textbf{Ablation Study on Verbal Annotations}.}
To investigate the impact of Verbal Annotation, we compared our proposed method against a variant trained with full document observations instead of Verbal Annotations, while keeping the same reranking strategy for both. 
As shown in Table \ref{table:ablation_planner}, the model trained with Verbal Annotations consistently and robustly outperforms the full-document counterpart across all evaluated benchmarks.
This performance gap underscores the critical role of Verbal Annotation in the RAG pipeline and validates that the inclusion of verbal feedback is essential for bridging the gap between retrieval and reasoning.

\noindent{\textbf{Verbal-R3's Context Utilization}.} 
We adopt analytical metrics in Section~\ref{sec:motivation} to study Verbal-R3. 
Table~\ref{table:anlysis_verbalr3} presents a comparative analysis of Search-R1 3B, Verbal-R3 3B, and an ablated version of Verbal-R3 3B that operates on the full document.
The results show that both Verbal-R3 and its ablation achieve higher retrieval accuracy than Search-R1, indicating that the performance gains primarily stem from the reranking component of the Verbal Reranker. 
Verbal-R3 consistently attains higher $\mathtt{CUE}$ scores independent of retrieval accuracy, suggesting that the performance gains largely stem from improved context utilization.


\begin{table}
\centering
\footnotesize
\resizebox{1\linewidth}{!}{
\begin{NiceTabular}{l|cccc|cc}
\toprule

Model & EM & F1 & RA-R & RA-L & $\mathtt{CUE}$-$\mathtt{R}$ & $\mathtt{CUE}$-$\mathtt{L}$ \\
\midrule

Search-R1
& 38.74 & 46.31 & 61.79 & 48.26 & 64.77 & 69.04 \\

Verbal-R3 \textit{w/o} VA
& 42.73 & 51.75 & 66.73 & \textbf{53.52} & 65.21 & 69.76 \\

Verbal-R3 & \textbf{45.36} & \textbf{54.66} & \textbf{69.06} & 52.86 & \textbf{67.54} & \textbf{71.64}  \\

\bottomrule

\end{NiceTabular}
}
\vspace{-0.3em}
\caption{Comparison of Verbal-R3, its ablation variant without Verbal Annotation, and Search-R1.}
\vspace{-5pt}
\label{table:anlysis_verbalr3}
\end{table}

\section{Conclusion}
This work unveiled the importance of Verbal Annotations, analytic narratives that serve as the logical bridge between a search query and retrieved contexts, in improving RAG.
Based on this finding, we proposed Verbal-R3, an agentic RAG framework consisting of a Generator that performs iterative retrieval and reasoning and a Verbal Reranker that produces Verbal Annotations to guide the Generator.
Verbal-R3 is complemented by novel relevance-guided test-time scaling that efficiently allocates test-time computation to promising reasoning trajectories. 
The success of Verbal-R3 at attaining state-of-the-art performance on complex QA benchmarks, along with thorough ablative studies, validate the contribution of each proposed technical component.
In summary, Verbal-R3 is an important step towards developing an interpretable and controllable agentic RAG framework that closes the gap between retrieval and LLM reasoning.

\section*{Limitations \& Potential Risks}

Our approach employs two LLM modules for RAG.
While we develop a lightweight reranker by distilling the outputs of a larger reasoning model, utilizing an additional LLM introduces additional computational overhead depending on the deployment setting. 
Although our Verbal Reranker is quite performant, it is still far from being perfect.
Therefore, the remaining errors present in the Verbal Annotations will accumulate during iterative retrieval and reasoning.
Due to practical constraints, our experiments were conducted using an offline retriever and publicly available datasets. 
Exploring the integration of a live web-search–based retrieval system is left for future study.

Though RAG reduces hallucination, the inherent probabilistic nature of LLMs inhibits a complete elimination of hallucinations.
In particular, a user can still receive inaccurate, misleading, or fabricated information when retrieved documents are incomplete, outdated, irrelevant, or incorrectly interpreted by the model. Additionally, errors may arise from improper retrieval ranking, ambiguous user queries, or overconfidence in generated responses. 
These risks highlight the necessity of complementary safeguards, such as confidence scoring, human oversight, and continuous evaluation of retrieval quality, especially in high-stakes domains. 
Addressing these limitations is essential to ensure the reliability of RAG-based systems.

\section*{Acknowledgments}

This work was supported by the Institute of Information \& Communications Technology Planning \& Evaluation (IITP) grants funded by the Korea government (MSIT) [NO.RS-2021-II211343, Artificial Intelligence Graduate
School Program (Seoul National University); No.2022-0-00959, RS-2022-II220959], by the National Research Foundation of Korea (NRF) grant [No.2022R1A3B1077720, 2022R1A5A7083908], BK21 FOUR Program of the Education and Research Program for Future ICT Pioneers, Seoul National University in 2026, and by the Samsung Electronics Co., Ltd [IO250520-12926-01].

\bibliography{custom}

\clearpage

\renewcommand{\theHsection}{A\arabic{section}}
\section*{Appendix}

\setcounter{section}{0}
\renewcommand\thesection{A\arabic{section}}
\setcounter{table}{0}
\renewcommand{\thetable}{A\arabic{table}}
\setcounter{figure}{0}
\renewcommand{\thefigure}{A\arabic{figure}}
\setcounter{equation}{0}
\renewcommand{\theequation}{A\arabic{equation}}

\section{Experimental Details}
\subsection{Baselines}
\label{sec:appendix baselines}
The baselines for rerankers are as follows:
\begin{itemize}
    \item MonoT5 \citep{nogueira2019document} uses a sequence-to-sequence model to evaluate a document–query pair. It is prompted with the query and the document, and then computes the logit of the output sequence.
    \item RankLLaMA \citep{ma2024fine} leverages a large language model to build a linear regressor.
    \item Rank1 \citep{weller2025rank1} uses test-time scaling with a large language model to produce more precise reranking.
\end{itemize}

The baselines for evaluation on QA benchmarks are as follows:

\begin{itemize}
    \item Direct Answer assesses the generator's internal knowledge without using any external context.
    \item Na\"ive RAG is a foundational baseline that appends retrieved documents in a single-step process.
    \item IRCoT \cite{ircot} uses a training-free iterative RAG framework.
    \item ITER-RETGEN \cite{iter-retgen} uses an iterative approach where the model's previous generation is utilized as context for retrieval.
    \item Search-R1 is a competitive baseline designed specifically for iterative RAG, optimizing multi-step search and reasoning via reinforcement learning.
\end{itemize}

The na\"ive majority voting baselines for test-time scaling method are as follows:
\begin{itemize}
    \item w/ unique query extraction  identifies identical queries within $N$ sampling trials and invokes the reranker only once per unique query. The resulting response is then replicated across all redundant instances.
    \item w/o unique query extraction executes a reranker call for every generated query, regardless of duplication.
\end{itemize}
While w/o unique query extraction approach incurs a higher number of reranker calls, it potentially increases response diversity by leveraging the stochasticity inherent in the Verbal Reranker's own sampling process.


\subsection{Training Details}
\label{sec:appendix Training Details}

\paragraph{Reranker Training.}
To train the Verbal Reranker, we utilize responses generated by GPT-OSS-120B on the Natural Questions (NQ) training set, which were subsequently filtered as described in the previous section. Specifically, we sampled 10,000 questions from NQ and retrieved the top 50 documents for each question using the E5 retriever, resulting in 500,000 question-document pairs. These pairs were then input as prompts to GPT-OSS-120B to generate Verbal Annotations. After applying our filtering procedure and splitting, 380,460 pairs (77\%) were selected for training. We trained Qwen2.5-Instruct-1.5B and Qwen2.5-Instruct-3B on this filtered dataset for 3 epochs using a batch size of 128.

\paragraph{Verbal-R3 Training.}
The training dataset is curated from the training sets of NQ and HotpotQA, and the Generators were trained for 2 epochs with a batch size of 128 and a temperature of 1.0. To ensure the model focuses on challenging reasoning paths, we evaluated each training sample from NQ and HotpotQA using the base Qwen2.5-3B model across 10 independent sampling trials; samples that were correctly answered in at least one trial were considered "trivial" and subsequently filtered out.
However, we exempted binary questions from this filtering to prevent the false exclusion of valid reasoning instances due to high random guessing probabilities.
\begin{table*}[t]
\centering
\footnotesize
\resizebox{1\linewidth}{!}{
\begin{NiceTabular}{l|ccccccccc|c}
\toprule
Model & ArguA & ClimF & DBP & FiQA & NFCorp & SciDoc & SciFact & Touche & TrecC & Avg \\
\midrule
BM25S         & 47.2 & 18.6 & 32.0 & 25.4 & 34.3 & 16.5 & 69.1 & 34.7 & 68.8 & 38.5 \\
\midrule
MonoT5-3B     & 42.5 & 25.4 & 44.5 & 46.5 & 37.8 & 19.3 & 76.1 & 30.7 & 79.6 & 44.7 \\
\midrule
RankLLaMA-7B  & 54.4 & 23.2 & 43.7 & 42.1 & 27.0 & 16.6 & 71.1 & 41.4 & 80.2 & 44.4 \\
RankLLaMA-13B & 49.3 & 24.5 & 44.9 & 44.1 & 28.1 & 18.1 & 72.7 & 39.2 & 80.8 & 44.6 \\
\midrule
Rank1-7B      & 42.8 & 15.0 & 38.9 & 39.5 & 36.2 & 17.2 & 77.2 & 22.8 & 81.9 & 41.3 \\
Rank1-14B     & 45.3 & 16.2 & 37.4 & 37.9 & 35.8 & 17.9 & 77.0 & 27.1 & 78.2 & 41.4 \\
Rank1-32B     & 57.6 & 15.8 & 40.7 & 41.8 & 36.9 & 19.6 & 76.8 & 19.9 & 81.9 & 43.4 \\
\midrule
Ours-1.5B     & 24.0 & 13.7 & 32.8 & 30.1 & 33.5 & 15.1 & 69.4 & 23.5 & 75.7 & 35.3 \\
Ours-3B       & 26.3 & 14.5 & 34.4 & 35.0 & 34.0 & 14.6 & 70.8 & 22.3 & 76.9 & 36.5 \\

\bottomrule
\end{NiceTabular}
}
\vspace{-0.3em}
\caption{The nDCG@10 performance of Verbal Reranker in BEIR benchmark. Baseline results originate from \citet{weller2025rank1}.}
\vspace{-5pt}
\label{table:retrieval_results}
\end{table*}

\begin{table*}
    \centering
    \resizebox{\textwidth}{!}{
        \begin{tabular}{lcccccccccccccc|cc}
        \toprule
        \multirow{2}{*}{\textbf{Method}} & \multicolumn{2}{c}{\textbf{2wiki}} & \multicolumn{2}{c}{\textbf{Bamboogle}} & \multicolumn{2}{c}{\textbf{HotpotQA}} & \multicolumn{2}{c}{\textbf{MuSiQue}} & \multicolumn{2}{c}{\textbf{NQ}} & \multicolumn{2}{c}{\textbf{PopQA}} & \multicolumn{2}{c|}{\textbf{TriviaQA}} & \multicolumn{2}{c}{\textbf{Average}} \\ \cmidrule(lr){2-3} \cmidrule(lr){4-5} \cmidrule(lr){6-7} \cmidrule(lr){8-9} \cmidrule(lr){10-11} \cmidrule(lr){12-13} \cmidrule(lr){14-15} \cmidrule(lr){16-17}
        & \textbf{EM} & \textbf{F1} & \textbf{EM} & \textbf{F1} & \textbf{EM} & \textbf{F1} & \textbf{EM} & \textbf{F1} & \textbf{EM} & \textbf{F1} & \textbf{EM} & \textbf{F1} & \textbf{EM} & \textbf{F1} & \textbf{EM} & \textbf{F1} \\ \midrule
        \textbf{Verbal-R3 3B} & \textbf{43.84} & \textbf{51.00} & \textbf{49.25} & \textbf{61.34} & \textbf{44.47} & \textbf{56.95} & \textbf{20.89} & \textbf{29.99} & \textbf{46.01} & \textbf{56.31} & \textbf{47.59} & \textbf{52.85} & \textbf{65.50} & \textbf{74.15} & \textbf{45.36} & \textbf{54.66} \\
        \textbf{w/o Verbal Annotation} & 39.09 & 45.02 & 43.55 & 56.57 & 43.74 & 55.79 & 20.07 & 28.37 & 45.12 & 54.97 & 44.96 & 50.17 & 62.61 & 71.39 & 42.73 & 51.75 \\ \bottomrule
        \end{tabular}
    }
    \vspace{-0.3em}
    \caption{Ablation study using Verbal-R3 3B.}
    \vspace{-5pt}
    \label{table:ablation_planner}
\end{table*}

For the Group Relative Policy Optimization (GRPO) objective, we configured the hyperparameters with $G=5$, $\epsilon=0.2$, and $\beta=0.001$.
Regarding the reward components, we assigned value of $R_{\mathrm{outcome}}=1.0$, $R_{\mathrm{format}}=0.2$, and $R_{\mathrm{answer}}=0.1$.
We employ 1.5B Verbal Reranker during the training phase to minimize computational overhead.
During training, we set the number of retrieved documents to $n=15$, from which the top $k=3$ reranker responses were integrated into the generator's context.


\subsection{Inference Details}
\label{sec:appendix Inference Details}
For all Verbal-R3 inference experiments, we employ sampling with a temperature of 1.0 and a top-$p$ of 0.95. The 3B Verbal Reranker was tasked with evaluating a candidate pool of $n=15$ retrieved documents. From this pool, the top $k=3$ reranked results were selected to be appended to the generation context for the Verbal-R3 Generator.
To generate outputs from the Verbal Reranker, we used a temperature of $0.6$ and top-$p = 0.95$. Inference was performed via the vLLM \cite{vllm} for optimized computational performance.

\section{Parsing Methods and Formats}
\label{sec:appendix Parsing Methods}
Following the protocol of Search-R1, we parse search queries only when they are enclosed within \texttt{<search>} and \texttt{</search>} tags.
Similarly, the final answer is extracted from the content within \texttt{<answer>} and \texttt{</answer>} tags.
A correctly formatted response is defined as a sequence of repeated reasoning segments and \texttt{<search> \{query\} </search>} blocks, concluding with a final \texttt{<answer> \{answer\} </answer>} tag.

\section{Evaluation on Reranking Benchmarks} \label{sec:appendix eval rerank}

Table \ref{table:retrieval_results} reports the performance of the Verbal Reranker on the BEIR benchmark. Following \cite{weller2025rank1}, all models rerank the top 100 documents initially retrieved by BM25S \cite{lu2024bm25s}. The results indicate that performance trends on BEIR do not align with the RAG results shown in Table \ref{table:main_reranker_rag}. This discrepancy suggests that existing reranking benchmarks may be inadequate for evaluating high-quality reranking modules, consistent with the observations of \citet{weller2025rank1}.

\begin{figure}[t]
  \centering
  \includegraphics[width=\columnwidth]{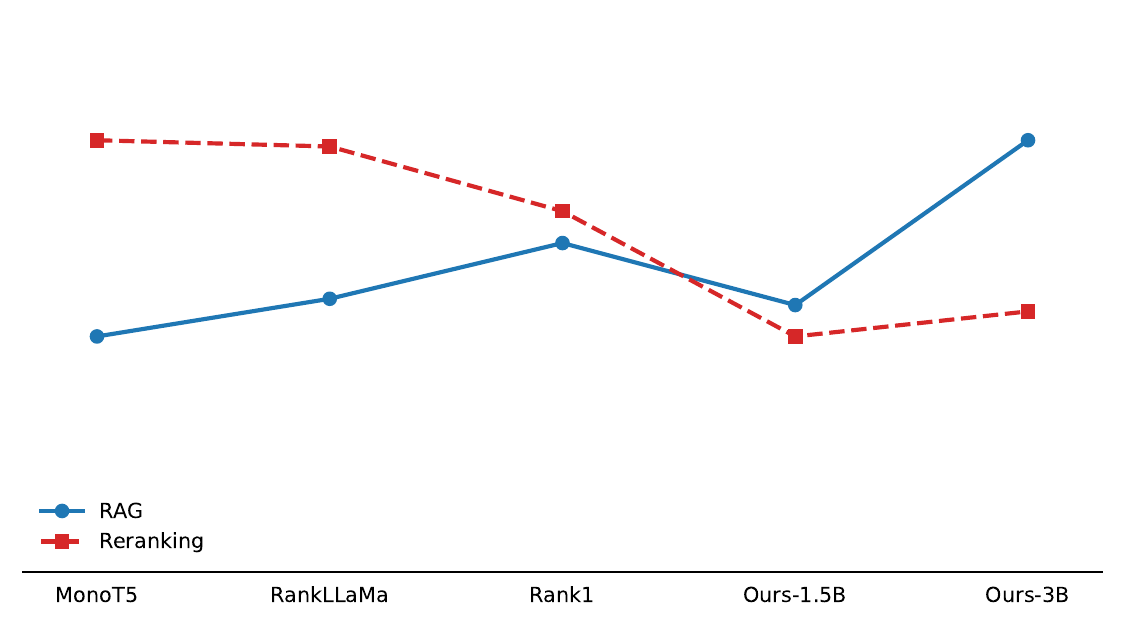}\\[-0.3em]
  \caption{\label{fig:rag_rerank} Performance of various rerankers on reranking benchmarks and in RAG settings. The divergent trends indicate that reranking benchmark scores do not reliably predict end-to-end RAG performance.}
\end{figure}

Figure \ref{fig:rag_rerank} illustrates the rerankers' performance on reranking benchmarks and in RAG settings. This further corroborates that BEIR benchmark scores do not reliably reflect a reranker's ultimate contribution to end-to-end RAG performance.

\section{Relevance-Guided Test-Time Scaling Algorithm}
The detailed algorithm for Relevance-Guided Test-Time Scaling is presented in Algorithm \ref{alg:tts}.
\setcounter{algorithm}{0} 
\renewcommand{\thealgorithm}{A\arabic{algorithm}}

\begin{algorithm*}[t]
\small
\caption{Relevance-Guided Test-Time Scaling}
\label{alg:tts}
\linespread{1.1}\selectfont
\begin{algorithmic}[1]
\Require Initial context $C$, total trajectory budget $N$, max turns $T$, scaling hyperparameter $\alpha$
\Ensure Final answer $A^*$

\State \textbf{Initialize:} $\mathcal{B}_0 \leftarrow \{(C, s=0, l=0)\}$, $\mathcal{F} \leftarrow \emptyset$, $t \leftarrow 0$

\While{$t < T$ \textbf{and} $|\mathcal{F}| < N$}
    \State $N_{e} \leftarrow N - |\mathcal{F}|$ \Comment{Calculate remaining budget}
    \State Sort branches $b_i = (C_i, s_i, l_i) \in \mathcal{B}_t$ in descending order of $s_i$, then $l_i$
    \State $n_{\mathrm{base}} \leftarrow \lfloor N_{e} / |\mathcal{B}_t| \rfloor, \quad \rho \leftarrow N_{e} \pmod{|\mathcal{B}_t|}$
    
    \ForAll{$b_i \in \mathcal{B}_t$} \Comment{Step 1: Generation}
        \State $m_i = n_{\mathrm{base}} + \mathbbm{1}(i < \rho)$
        \State Generate $m_i$ responses $\{y_{i,j}\}_{j=1}^{m_i}$ with context $C_i$
        \State $\mathcal{M}_i \leftarrow \emptyset$ \Comment{Initialize query-response mapping}
        \ForAll{$y_{i,j}$}
            \If{$y_{i,j}$ contains answer $a_{i,j}$}
                \State $\mathcal{F} \leftarrow \mathcal{F} \cup \{a_{i,j}\}$
            \ElsIf{$y_{i,j}$ contains query $q_{i,j}$}
                \State Collect $q_{i,j}$ for reranking
                \State $\mathcal{M}_i(q_{i,j}) \leftarrow y_{i,j}$
            \EndIf
        \EndFor
    \EndFor

    \State $\mathcal{V} \leftarrow \text{Unique}(\text{all collected } q_{i,j})$ \Comment{Step 2: Reranking}
    \ForAll{$v \in \mathcal{V}$}
        \State $\mathcal{S}_v \leftarrow \text{Rerank}(v)$, $s_{\max, v} \leftarrow \max(\mathcal{S}_v)$
        \State $l_v \leftarrow \mathrm{GetRerankerLogit}(s_{\max, v})$
    \EndFor

    \State $\mathcal{B}_{t+1} \leftarrow \emptyset$
    \ForAll{$b_i \in \mathcal{B}_t$} \Comment{Step 3: Branch-wise Sampling}
        \State $\mathcal{U}_i \leftarrow \text{Unique}(\{q_{i,j} \in \text{responses of } b_i\})$
        \State $s_{\mathrm{best}, b_i} \leftarrow \max_{q \in \mathcal{U}_i} \{s_{\max, q}\}$
        \ForAll{$q \in \mathcal{U}_i$}
            \State $P(q) \leftarrow (\frac{s_{\max, q}}{s_{\mathrm{best}, b_i}})^\alpha$
            \State Sample $z \sim \text{Bernoulli}(P(q))$
            \If{$z = 1$}
                \State $y \leftarrow \mathcal{M}_i(q)$, $C' \leftarrow C_i \oplus y \oplus \text{Obs}_q$
                \State $b_{\mathrm{new}} \leftarrow (C', s_{\max, q}, l_q)$ 
                \State $\mathcal{B}_{t+1} \leftarrow \mathcal{B}_{t+1} \cup \{b_{\mathrm{new}}\}$
            \EndIf
        \EndFor
    \EndFor
    \State $t \leftarrow t + 1$
\EndWhile
\State \Return $A^* = \text{MajorityVote}(\mathcal{F})$
\end{algorithmic}
\end{algorithm*}

\section{Ablation Study on Verbal Annotations}
The results of ablation study on Verbal Annotations are shown in Table \ref{table:ablation_planner}.


\begin{table*}[!t]
    \centering
    \resizebox{\textwidth}{!}{
        \begin{tabular}{lcccccccccccccc|cc|c}
        \toprule
        \multirow{2}{*}{\textbf{Method}} 
        & \multicolumn{2}{c}{\textbf{2Wiki}} 
        & \multicolumn{2}{c}{\textbf{Bamboogle}} 
        & \multicolumn{2}{c}{\textbf{HotpotQA}} 
        & \multicolumn{2}{c}{\textbf{MuSiQue}} 
        & \multicolumn{2}{c}{\textbf{NQ}} 
        & \multicolumn{2}{c}{\textbf{PopQA}} 
        & \multicolumn{2}{c|}{\textbf{TriviaQA}} 
        & \multicolumn{2}{c|}{\textbf{Average}}
        & \multirow{2}{*}[-0.4em]{\shortstack{\textbf{FLOPs} \\ \textbf{($\times 10^{15}$)}}} \\
        \cmidrule(lr){2-3} \cmidrule(lr){4-5} \cmidrule(lr){6-7} \cmidrule(lr){8-9}
        \cmidrule(lr){10-11} \cmidrule(lr){12-13} \cmidrule(lr){14-15} \cmidrule(lr){16-17}
        & \textbf{EM} & \textbf{F1}
        & \textbf{EM} & \textbf{F1}
        & \textbf{EM} & \textbf{F1}
        & \textbf{EM} & \textbf{F1}
        & \textbf{EM} & \textbf{F1}
        & \textbf{EM} & \textbf{F1}
        & \textbf{EM} & \textbf{F1}
        & \textbf{EM} & \textbf{F1}
        & \\
        \midrule
        \textbf{w/o Reranker}
        & 35.7 & 42.2
        & 39.8 & 53.0
        & 40.4 & 52.3
        & 17.8 & 26.1
        & \textbf{40.3} & \textbf{52.8}
        & 44.1 & 51.0
        & 66.4 & 74.7
        & 40.6 & 50.3
        & 0.207 \\
        
        \textbf{w/ 3B Verbal Reranker}
        & \textbf{37.3} & \textbf{43.6}
        & \textbf{47.4} & \textbf{59.5}
        & \textbf{41.1} & \textbf{53.2}
        & \textbf{18.3} & \textbf{27.0}
        & \textbf{40.3} & 52.6
        & \textbf{44.2} & \textbf{51.6}
        & \textbf{67.2} & \textbf{75.6}
        & \textbf{42.3} & \textbf{51.9}
        & 0.236 \\
        \bottomrule
        \end{tabular}
    }
    \vspace{-0.3em}
    \caption{Evaluation in the iterative retrieval setting with Qwen2.5-32B-Instruct as the Generator, with and without our 3B Verbal Reranker.}
    \label{table:scaling_32B_iterative}
\end{table*}

\section{Analysis of Scaling and Computational Cost}
\label{sec:appendix scaling}
To prioritize research efficiency, we restrict RL-based generator training to the 3B and 7B models.
To further analyze generator scaling and the computational cost of the Verbal Reranker, we additionally evaluate Qwen2.5-32B-Instruct model as the Generator on the RAG benchmarks and report the corresponding inference cost in Table~\ref{table:scaling_32B_iterative}.
Under the iterative retrieval setting, the results show that our Verbal Reranker consistently improves both EM and F1 even when paired with a 32B Generator, demonstrating that the benefits of our Verbal Reranker transfer robustly beyond the model scales.

From the perspective of computational efficiency, incorporating the Verbal Reranker increases the average F1 score by 3.13\% while introducing only 13.8\% additional FLOPs.
In contrast, scaling Search-R1 from a 3B to a 7B backbone yields an 8.18\% F1 improvement at the cost of approximately 133\% more FLOPs.
This comparison highlights that our approach provides a more favorable performance-cost trade-off.
Moreover, because the Verbal Reranker remains a relatively small external module, its relative computational overhead becomes less significant as the Generator size increases.

\begin{table*}[!t]
    \centering
    \resizebox{\textwidth}{!}{
        \begin{tabular}{lccccccc|c}
        \toprule
        \textbf{Method} 
        & \textbf{2Wiki}
        & \textbf{Bamboogle}
        & \textbf{HotpotQA}
        & \textbf{MuSiQue}
        & \textbf{NQ}
        & \textbf{PopQA}
        & \textbf{TriviaQA}
        & \textbf{Average} \\
        \midrule
        \multicolumn{9}{l}{\textit{\textbf{Qwen2.5-3B}}} \\
        AFM & 43.4 & 45.6 & 41.1 & 19.0 & 39.3 & 42.4 & 58.2 & 41.3 \\
        CoDA & 40.8 & 36.8 & 40.7 & 18.7 & 44.1 & 45.2 & 60.8 & 41.0 \\
        Thinker & \textbf{46.9} & 42.4 & 40.0 & \textbf{21.4} & 43.9 & 46.9 & 59.8 & 43.0 \\
        InfoFlow & 45.2 & 41.2 & \textbf{44.6} & 21.0 & 44.5 & 47.0 & 63.7 & 43.9 \\
        Verbal-R3 (Ours) & 43.8 & \textbf{49.3} & 44.5 & 20.9 & \textbf{46.0} & \textbf{47.6} & \textbf{65.5} & \textbf{45.4} \\
        \midrule
        \multicolumn{9}{l}{\textit{\textbf{Qwen2.5-7B}}} \\
        AFM & \textbf{49.2} & 49.6 & 43.9 & 22.3 & 43.9 & 46.5 & 63.3 & 45.5 \\
        Thinker & 46.9 & 48.0 & 42.1 & 22.1 & 45.0 & \textbf{48.4} & 64.2 & 45.2 \\
        InfoFlow & 47.2 & 47.6 & 44.3 & 21.9 & 47.2 & 48.1 & \textbf{68.1} & 46.3 \\
        Verbal-R3 (Ours) & 47.2 & \textbf{53.6} & \textbf{47.5} & \textbf{25.2} & \textbf{48.3} & 48.3 & \textbf{68.1} & \textbf{48.3} \\
        \bottomrule
        \end{tabular}
    }
    \vspace{-0.3em}
    \caption{Comparison with agentic RAG frameworks under 3B and 7B backbone settings. AFM reports performance using an LLM-as-a-Judge metric, whereas the remaining methods are evaluated with Exact Match.}
    \vspace{-1.3em}
    \label{table:agentic_rag_comparison}
\end{table*}

\section{Comparisons with Advanced Agentic RAG Frameworks}
\label{sec:appendix Comparison Agentic}
In addition to Search-R1, which is used as a baseline in the main paper, we further compare our method against advanced agentic RAG frameworks.
Specifically, we include AFM~\cite{AFMli2025chainofagentsendtoendagentfoundation}, CoDA~\cite{CoDA}, Thinker~\cite{thinker}, and InfoFlow~\cite{infoflowluo2025infoflowreinforcingsearchagent} as additional baselines. The comparison results are presented in Table~\ref{table:agentic_rag_comparison}. For these baselines, we report the performance numbers as provided in their original papers.

\section{Use of AI Writing Assistant}
AI-driven writing assistant tools, such as ChatGPT, Gemini, and Claude, were used to paraphrase sentences for improved clarity and to search for synonyms or similar phrases.
Beyond this, no additional AI assistance was involved during the research process or paper write-up.


\begin{figure*}[t]
  \centering
  \includegraphics[width=\linewidth]{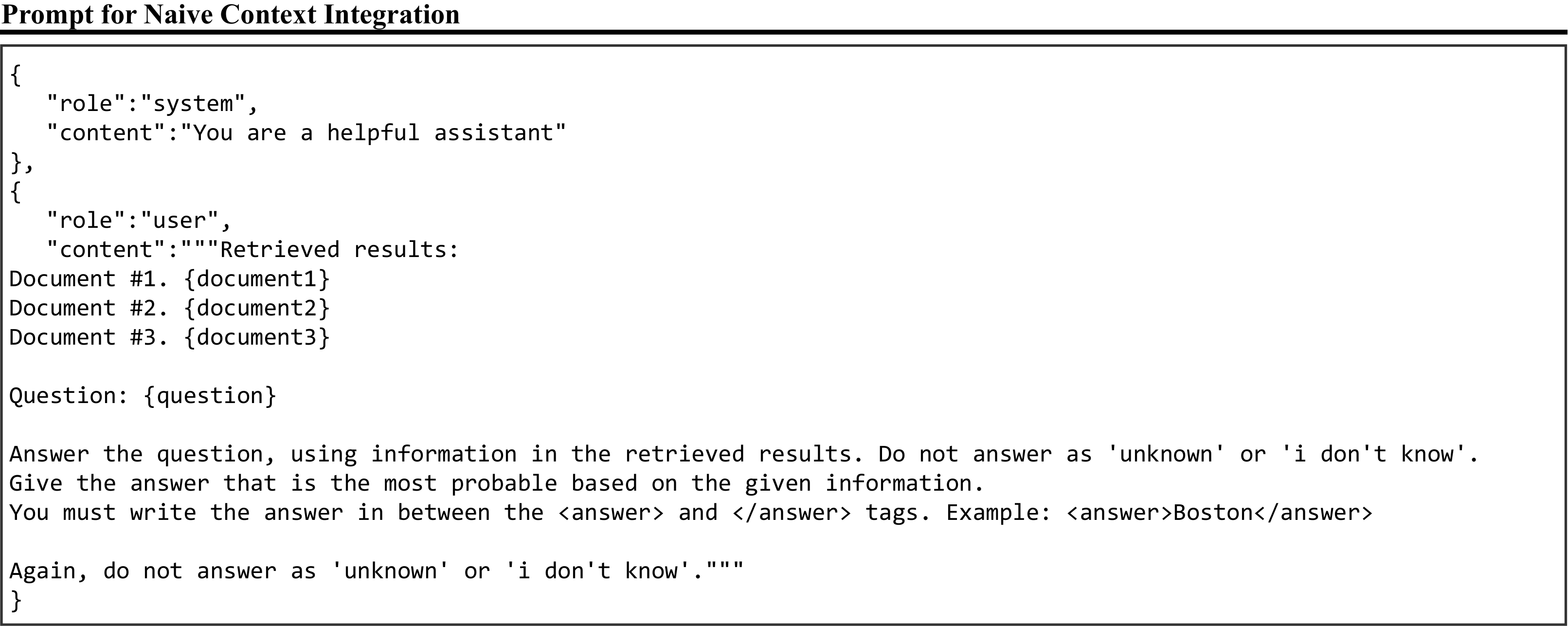}\\[-0.3em]
  \caption{\label{fig:p_naive_rag} Prompt template for na\"ive context integration.}
\end{figure*}

\begin{figure*}[t]
  \centering
  \includegraphics[width=\linewidth]{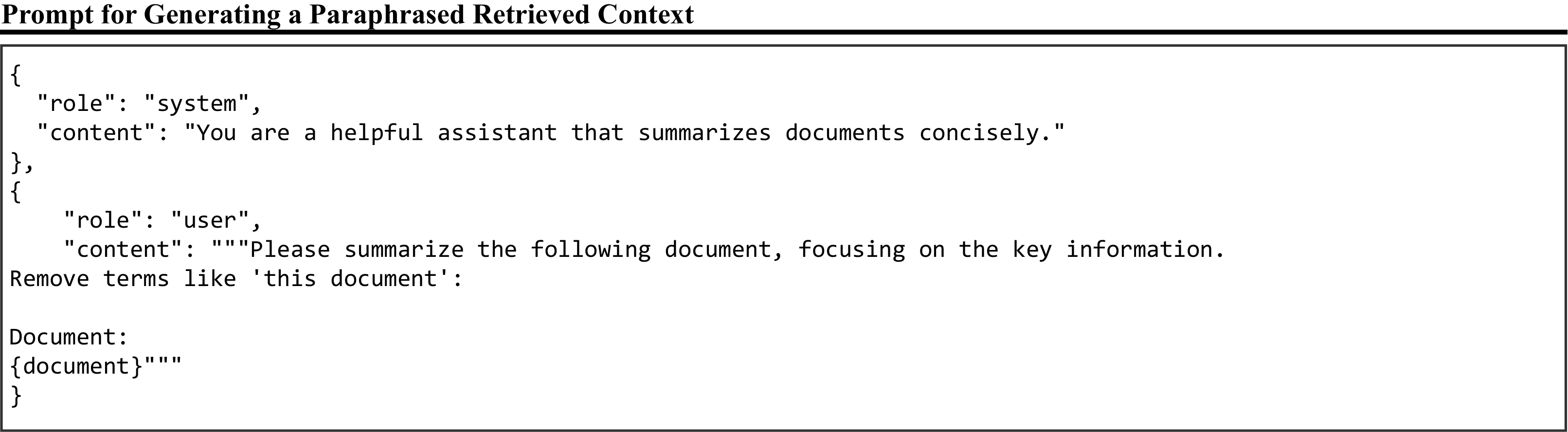}\\[-0.3em]
  \caption{\label{fig:p_paraphrase} Prompt template for generating paraphrased contexts.}
\end{figure*}

\begin{figure*}[t]
  \centering
  \includegraphics[width=\linewidth]{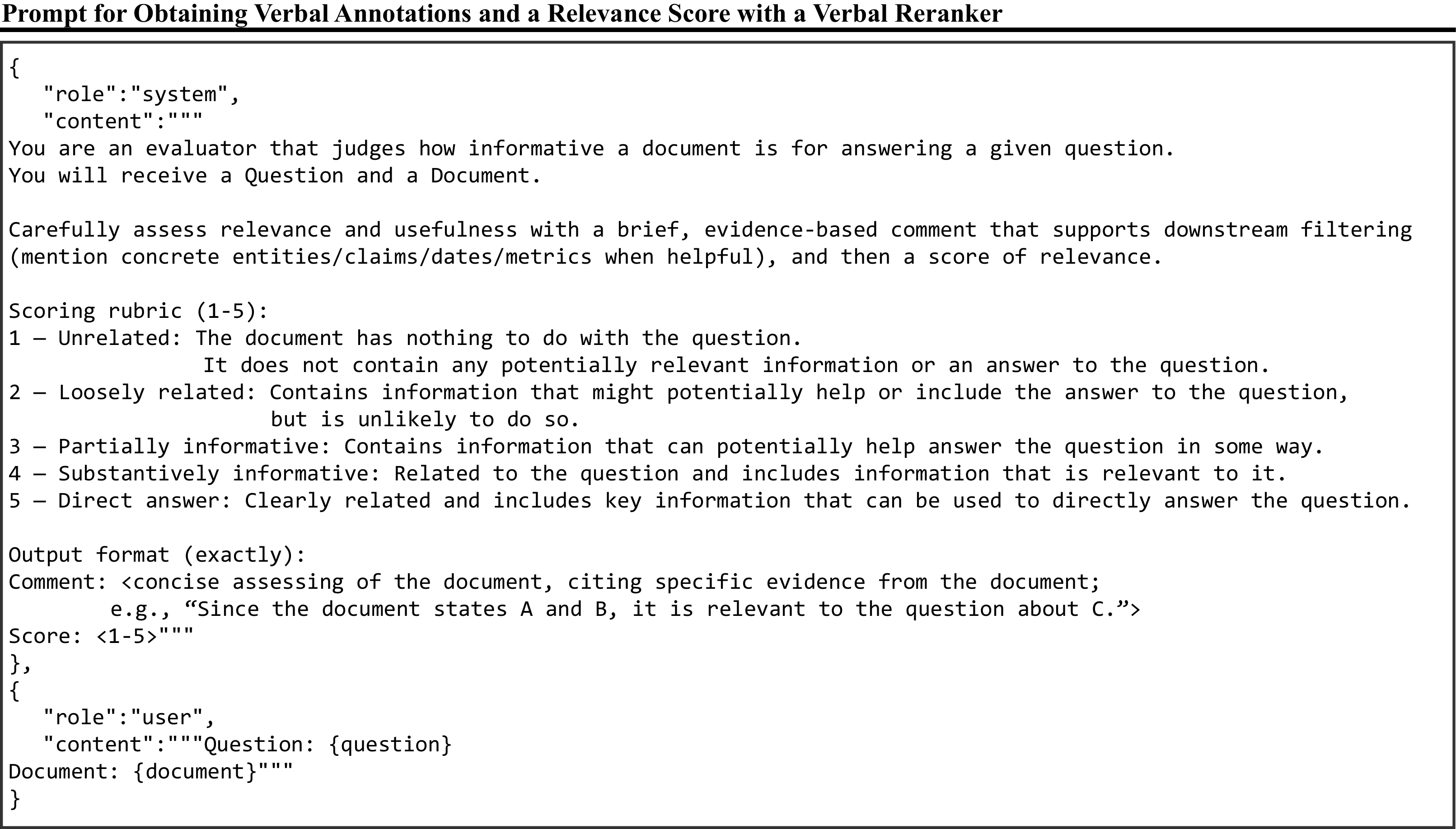}\\[-0.3em]
  \caption{\label{fig:p_verbal_reranker} Prompt template for generating Verbal Annotations and a relevance score with a Verbal Reranker.}
\end{figure*}

\begin{figure*}[t]
  \centering
  \includegraphics[width=\linewidth]{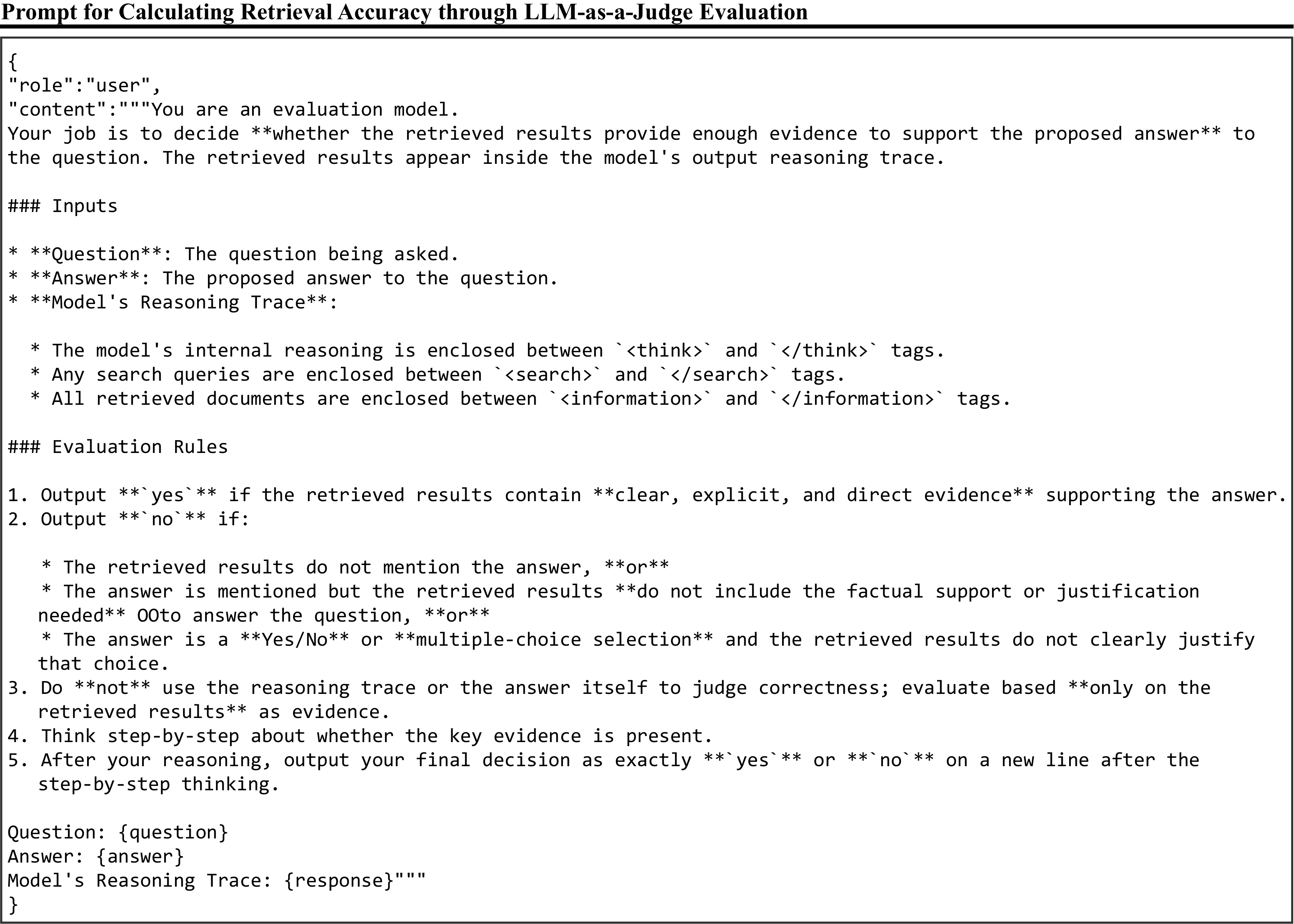}\\[-0.3em]
  \caption{\label{fig:p_cue_l} Prompt template for evaluating retrieval accuracy through LLM-as-a-Judge evaluation.}
\end{figure*}

\begin{figure*}[t]
  \centering
  \includegraphics[width=\linewidth]{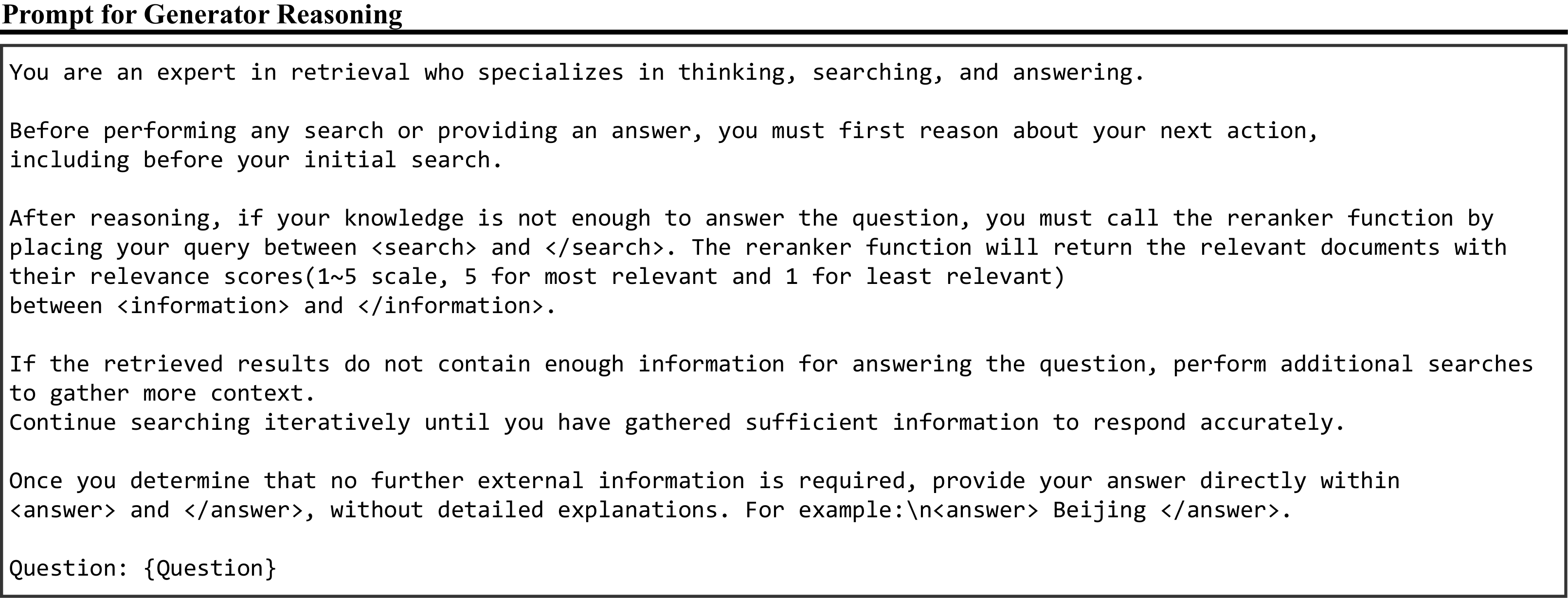}\\[-0.3em]
  \caption{\label{fig:p_generator} Prompt template for iterative retrieval and reasoning with the Generator.}
\end{figure*}

\end{document}